\pdfoutput=1

\documentclass[11pt]{article}

\usepackage{naacl2021}

\usepackage{times}
\usepackage{latexsym}

\usepackage{graphicx}
\usepackage{tikz}
\usepackage{amsmath}
\usepackage{amssymb}
\usepackage{booktabs}
\usepackage{color}
\usepackage{hyperref}
\usepackage{multirow}
\usepackage{subfigure}
\usepackage{graphicx}
\usepackage{textcomp}
\usepackage{amssymb}
\usepackage{microtype}
\usepackage{url}

\newcommand\blfootnote[1]{%
  \begingroup
  \renewcommand\thefootnote{}\footnote{#1}%
  \addtocounter{footnote}{-1}%
  \endgroup
}

\usepackage{enumitem}

\usepackage{multirow}

\usepackage[T1]{fontenc}

\usepackage[utf8]{inputenc}

\usepackage{microtype}

\title{Multimodal End-to-End Sparse Model for Emotion Recognition}
\author{Wenliang Dai*, Samuel Cahyawijaya*, Zihan Liu, Pascale Fung \\
Center for Artificial Intelligence Research (CAiRE)\\
Department of Electronic and Computer Engineering\\
The Hong Kong University of Science and Technology, Clear Water Bay, Hong Kong\\
\texttt{\{wdaiai,scahyawijaya,zliucr\}@connect.ust.hk, pascale@ece.ust.hk}}

\begin{document}
\maketitle
\begin{abstract}
Existing works on multimodal affective computing tasks, such as emotion recognition, generally adopt a two-phase pipeline, first extracting feature representations for each single modality with hand-crafted algorithms and then performing end-to-end learning with the extracted features.
However, the extracted features are fixed and cannot be further fine-tuned on different target tasks, and manually finding feature extraction algorithms does not generalize or scale well to different tasks, which can lead to sub-optimal performance. 
In this paper, we develop a fully end-to-end model that connects the two phases and optimizes them jointly. In addition, we restructure the current datasets to enable the fully end-to-end training.
Furthermore, to reduce the computational overhead brought by the end-to-end model, we introduce a sparse cross-modal attention mechanism for the feature extraction.
Experimental results show that our fully end-to-end model significantly surpasses the current state-of-the-art models based on the two-phase pipeline. Moreover, by adding the sparse cross-modal attention, our model can maintain performance with around half the computation in the feature extraction part. \blfootnote{* Equal contribution.}
\blfootnote{Code is available at: \url{https://github.com/wenliangdai/Multimodal-End2end-Sparse}}
\end{abstract}

\section{Introduction} \label{intro}
Humans show their characteristics through not only the words they use, but also the way they speak and their facial expressions. Therefore, in multimodal affective computing tasks, such as emotion recognition, there are usually three modalities: textual, acoustic, and visual. 
One of the main challenges in these tasks is how to model the interactions between different modalities, as they contain both supplementary and complementary information~\citep{baltruvsaitis2018multimodal}. 

In the existing works, we discover that a two-phase pipeline is generally used~\citep{zadeh2018memory,zadeh2018multimodal,tsai2018learning,tsai2019multimodal,rahman2020integrating}. In the first phase, given raw input data, feature representations are extracted with hand-crafted algorithms for each modality separately, while in the second phase, end-to-end multimodal learning is performed using extracted features. 
However, there are three major defects of this two-phase pipeline: 1) the features are fixed after extraction and cannot be further fine-tuned on target tasks; 2) manually searching for appropriate feature extraction algorithms is needed for different target tasks; and 3) the hand-crafted model considers very few data points to represent higher-level feature, which might not capture all the useful information.
These defects can result in sub-optimal performance.


\begin{figure}[t]
    \centering
    \includegraphics[width=\linewidth]{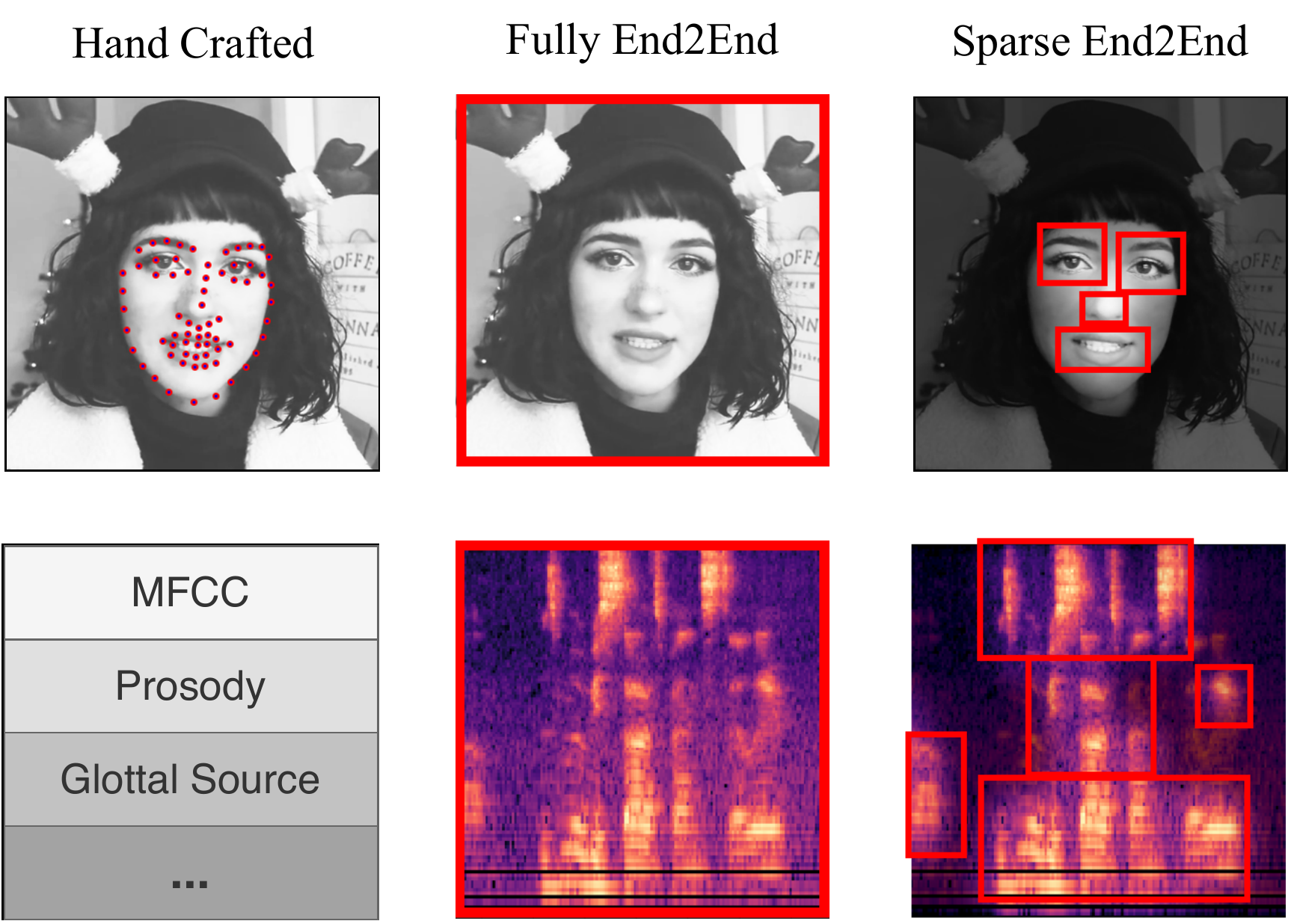}
    \caption{An illustration of feature extraction from hand-crafted model (\textit{left}), fully end-to-end model (\textit{middle}), and sparse end-to-end model (\textit{right}). The red dots represent the keypoints extracted by hand-crafted models. The areas formed by red lines represent the regions of interest that are processed by (sparse) end-to-end models to extract the features.}
    \label{fig:intro}
\end{figure}

In this paper, we propose a fully end-to-end model that connects the two phases together and optimizes them jointly. In other words, the model receives raw input data and produces the output predictions, which allows the features to be learned automatically through the end-to-end training.
However, the current datasets for multimodal emotion recognition cannot be directly used for the fully end-to-end training, and we thus conduct a data restructuring to make this training possible.
The benefits from the end-to-end training are that the features are optimized on specific target tasks, and there is no need to manually select feature extraction algorithms. 
Despite the advantages of the end-to-end training, it does bring more computational overhead compared to the two-phase pipeline, and exhaustively processing all the data points makes it computationally expensive and prone to over-fitting. 
Thus, to mitigate these side-effects, we also propose a multimodal end-to-end sparse model, a combination of a sparse cross-modal attention mechanism and sparse Convolutional Neural Network (CNN)~\cite{SubmanifoldSparseConvNet}, to select the most relevant features for the task and reduce the redundant information and noise in the video and audio.

Experimental results show that the simply end-to-end training model is able to consistently outperform the existing state-of-the-art models which are based on the two-phase pipeline. Moreover, the incorporation of the sparse cross-modal attention and sparse CNN is able to greatly reduce the computational cost and maintain the performance.

We summarize our contributions as follows.
\begin{itemize}
    \item To the best of our knowledge, we are the first to apply a fully end-to-end trainable model for the multimodal emotion recognition task.
    \item We restructure the existing multimodal emotion recognition datasets to enable the end-to-end training and cross-modal attention based on the raw data.
    \item We show that the fully end-to-end training significantly outperforms the current state-of-the-art two-phase models, and the proposed sparse model can greatly reduce the computational overhead while maintaining the performance of the end-to-end training. We also conduct a thorough analysis and case study to improve the interpretability of our method.
\end{itemize}

\section{Related Works} \label{related_works}
Human affect recognition is a popular and widely studied research topic~\citep{Mirsamadi2017AutomaticSE,Zhang2017SentimentAA,Xu2020EmoGraphCE,Dai2020KungfupandaAS}. In recent years, there is a trend to leverage multimodal information to tackle these research tasks, such as emotion recognition~\citep{busso2008iemocap}, sentiment analysis~\citep{zadeh2016mosi,zadeh2018multimodal}, personality trait recognition~\citep{pom}, etc, have drawn more and more attention. 
Different methods have been proposed to improve the performance and cross-modal interactions. 
In earlier works, early fusion~\citep{morency2011towards,perez2013utterance} and late fusion~\citep{zadeh2016mosi,wang2017select} of modalities were widely adopted. Later, more complex approaches were proposed. For example, \citet{zadeh2017tensor} introduced the Tensor Fusion Network to model the interactions of the three modalities by performing the Cartesian product, while \citep{wang2019words} used an attention gate to shift the words using the visual and acoustic features. In addition, based on the Transformer~\citep{vaswani2017attention}, \citet{tsai2019multimodal} introduced the Multimodal Transformer to improve the performance given unaligned multimodal data, and \citet{rahman2020integrating} introduced a multimodal adaptation gate to integrate visual and acoustic information into a large pre-trained language model. However, unlike some other multi-modal tasks~\citep{Chen2017SCACNNSA,Yu2019DeepMC,Li2019VisualBERTAS} using fully end-to-end learning, all of these methods require a feature extraction phase using hand-crafted algorithms (details in Section~\ref{baseline}), which makes the whole approach a two-phase pipeline.

\section{Dataset Reorganization}
The fully end-to-end multimodal model requires the inputs to be raw data for the three modalities (visual, textual and acoustic).
The existing multimodal emotion recognition datasets cannot be directly applied for the fully end-to-end training for two main reasons.
First, the datasets provide split of training, validation and test data for the hand-crafted features as the input of the model and emotion or sentiment labels as the output of the model. However, this dataset split cannot be directly mapped to the raw data since the split indices cannot be matched back to the raw data.
Second, the labels of the data samples are aligned with the text modality. However, the visual and acoustic modalities are not aligned with the textual modality in the raw data, which disables the fully end-to-end training.
To make the existing datasets usable for the fully end-to-end training and evaluation, we need to reorganize them according to two steps: 1) align the text, visual and acoustic modalities; 2) split the aligned data into training, validation and test sets.


In this work we reorganize two emotion recognition datasets: Interactive Emotional Dyadic Motion Capture (IEMOCAP) and CMU Multimodal Opinion Sentiment and Emotion Intensity (CMU-MOSEI). Both have multi-class and multi-labelled data for multimodal emotion recognition obtained by generating raw utterance-level data, aligning the three modalities, and creating a new split over the aligned data.
In the following section, we will first introduce the existing datasets, and then we will give a detailed description of how we reorganize them.

\subsection{IEMOCAP}
IEMOCAP~\cite{busso2008iemocap} is a multimodal emotion recognition dataset containing 151 videos. In each video, two professional actors conduct dyadic conversations in English. The dataset is labelled by nine emotion categories, but due to the data imbalance issue, we take the six main categories: \textit{angry}, \textit{happy}, \textit{excited}, \textit{sad}, \textit{frustrated}, and \textit{neutral}. As the dialogues are annotated at the utterance level, we clip the data per utterance from the provided text transcription time, which results in 7,380 data samples in total. Each data sample consists of three modalities: audio data with a sampling rate of 16 kHz, a text transcript, and image frames sampled from the video at 30 Hz. The provided pre-processed data from the existing work~\cite{busso2008iemocap} \footnote{\url{http://immortal.multicomp.cs.cmu.edu/raw_datasets/processed_data/iemocap}} doesn't provide an identifier for each data sample, which makes it impossible to reproduce it from the raw data. To cope with this problem, we create a new split for the dataset by randomly allocating 70\%, 10\%, and 20\% of data into the training, validation, and testing sets, respectively. The statistics of our dataset split are shown in Table~\ref{tab:iemocap-dataset-statistics}.

\begin{table}[!t]
\centering
\resizebox{0.49\textwidth}{!}{
\begin{tabular}{l|cc|ccc}
\toprule
\textbf{Label} &
\textbf{\begin{tabular}[c]{@{}c@{}}Avg. word \\ length\end{tabular}} &\textbf{\begin{tabular}[c]{@{}c@{}}Avg. clip \\ duration (s)\end{tabular}} & \textbf{\begin{tabular}[c]{@{}c@{}}Train \\ size\end{tabular}} & \textbf{\begin{tabular}[c]{@{}c@{}}Valid \\ size\end{tabular}} & \textbf{\begin{tabular}[c]{@{}c@{}}Test \\ size\end{tabular}} \\
\midrule
Anger & 15.96 & 4.51 & 757 & 112 & 234 \\
Excited & 16.79 & 4.78 & 736 & 92 & 213 \\
Frustrated & 17.14 & 4.71 & 1298 & 180 & 371 \\
Happiness & 13.58 & 4.34 & 398 & 62 & 135 \\
Neutral & 13.08 & 3.90 & 1214 & 173 & 321 \\
Sadness & 14.82 & 5.50 & 759 & 118 & 207 \\
\bottomrule
\end{tabular}
}
\caption{Statistics of our IEMOCAP dataset split.}
\label{tab:iemocap-dataset-statistics}
\end{table}



\subsection{CMU-MOSEI} 
CMU-MOSEI~\cite{zadeh2018multimodal} comprises 3,837 videos from 1,000 diverse speakers with six emotion categories: \textit{happy}, \textit{sad}, \textit{angry}, \textit{fearful}, \textit{disgusted}, and \textit{surprised}. It is annotated at utterance-level, with a total of 23,259 samples.  Each data sample in CMU-MOSEI consists of three modalities: audio data with a sampling rate of 44.1 kHz, a text transcript, and image frames sampled from the video at 30 Hz. We generate the utterance-level data from the publicly accesible raw CMU-MOSEI dataset. \footnote{\url{http://immortal.multicomp.cs.cmu.edu/raw_datasets/processed_data/cmu-mosei/seq_length_20/}} The generated utterances are perfectly matched with the preprocessed data from the existing work~\citep{zadeh2018multimodal}, but there are two issues with the existing dataset: 1) in includes many misaligned data samples; and 2) many of the samples do not exist in the generated data, and vice versa, in the provided standard split from the CMU MultiModal SDK. \footnote{\url{https://github.com/A2Zadeh/CMU-MultimodalSDK}} To cope with the first issue, we perform data cleaning to remove the misaligned samples, which results in 20,477 clips in total. We then create a new dataset split following the CMU-MOSEI split for the sentiment classification task.~\footnote{\url{http://immortal.multicomp.cs.cmu.edu/raw_datasets/processed_data/cmu-mosei/seq_length_50/mosei_senti_data_noalign.pkl}} The statistics of the new dataset split setting are shown in Table~\ref{tab:mosei-dataset-statistics}.

\begin{table}[!t]
\centering
\resizebox{0.49\textwidth}{!}{
\begin{tabular}{l|cc|ccc}
\toprule
\textbf{Label} &
\textbf{\begin{tabular}[c]{@{}c@{}}Avg. word \\ length\end{tabular}} &\textbf{\begin{tabular}[c]{@{}c@{}}Avg. clip \\ duration (s)\end{tabular}} & \textbf{\begin{tabular}[c]{@{}c@{}}Train \\ size\end{tabular}} & \textbf{\begin{tabular}[c]{@{}c@{}}Valid \\ size\end{tabular}} & \textbf{\begin{tabular}[c]{@{}c@{}}Test \\ size\end{tabular}} \\
\midrule
Anger & 7.75 & 23.24 & 3267 & 318 & 1015 \\
Disgust & 7.57 & 23.54 & 2738 & 273 & 744 \\
Fear & 10.04 & 28.82 & 1263 & 169 & 371 \\
Happiness & 8.14 & 24.12 & 7587 & 945 & 2220 \\
Sadness & 8.12 & 24.07 & 4026 & 509 & 1066 \\
Surprise & 8.40 & 25.95 & 1465 & 197 & 393 \\
\bottomrule
\end{tabular}
}
\caption{Statistics of our CMU-MOSEI dataset split.}
\label{tab:mosei-dataset-statistics}
\end{table}


\begin{figure*}[t]
    \centering
    \includegraphics[width=\linewidth]{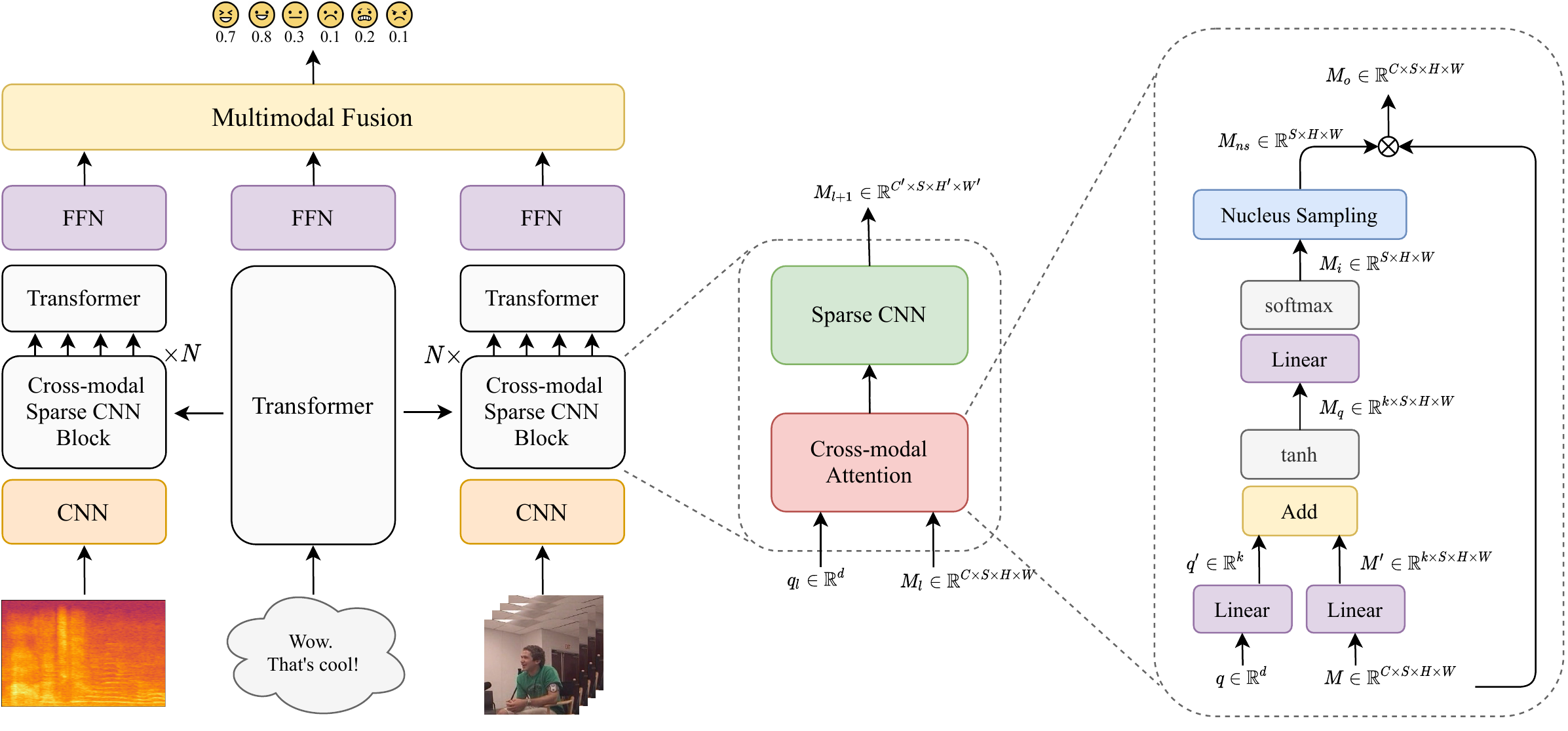}
    \caption{Architecture of our Multimodal End-to-end Sparse Model (MESM). On the left, we show the general architecture flow. In the middle and on the right, we exhibit the details of the cross-modal sparse CNN block, especially the cross-modal attention layer, which is the key to making the CNN model sparse.}
    \label{fig:main_model}
\end{figure*}

\section{Methodology} \label{methodology}

\subsection{Problem Definition}
We define $I$ multimodal data samples as $X=\{(t_i,a_i,v_i)\}^I_{i=1}$, in which $t_i$ is a sequence of words, $a_i$ is a sequence of spectrogram chunks from the audio, and $v_i$ is a sequence of RGB image frames from the video. $Y=\{y_i\}^I_{i=1}$ denotes the annotation for each data sample.

\subsection{Fully End-to-End Multimodal Modeling} \label{fe2e}
We build a fully end-to-end model which jointly optimizes the two separate phases (feature extraction and multimodal modelling).

For each spectrogram chunk and image frame in the visual and acoustic modalities, we first use a pre-trained CNN model (an 11-layer VGG~\citep{simonyan2014very} model) to extract the input features, which are then flattened to vector representations using a linear transformation.
After that, we can obtain a sequence of representations for both visual and acoustic modalities. Then, we use a Transformer~\citep{vaswani2017attention} model to encode the sequential representations since it contains positional embeddings to model the temporal information.
Finally, we take the output vector at the ``CLS'' token and apply a feed-forward network (FFN) to get the classification scores. 

In addition, to reduce GPU memory and align with the two-phase baselines which extract visual features from human faces, we use a MTCNN~\citep{mtcnn} model to get the location of faces for the image frames before feeding them into the VGG. For the textual modality, the Transformer model is directly used to process the sequence of words. Similar to the visual and acoustic modalities, we consider the feature at the ``CLS'' token as the output feature and feed it into a FFN to generate the classification scores. We take a weighted sum of the classification scores from each modality to make the final prediction score.

\subsection{Multimodal End-to-end Sparse Model}
Although the fully end-to-end model has many advantages over the two-phase pipeline, it also brings much computational overhead. To reduce this overhead without downgrading the performance, we introduce our Multimodal End-to-end Sparse Model (MESM).
Figure~\ref{fig:main_model} shows the overall architecture of MESM. In contrast to the fully end-to-end model, we replace the original CNN layers (except the first one for low-level feature capturing) with $N$ cross-modal sparse CNN blocks. A cross-modal sparse CNN block consists of two parts, a cross-modal attention layer and a sparse CNN model that contains two sparse VGG layers and one sparse max-pooling layer.

\subsubsection{Cross-modal Attention Layer}
The cross-modal attention layer accepts two inputs: a query vector $q \in \mathbb{R}^d$ and a stack of feature maps $M \in \mathbb{R}^{C \times S \times H \times W}$, where $C,S,H,$ and $W$ are the number of channels, sequence length, height, and width, respectively. Then, the cross-modal spatial attention is performed over the feature maps using the query vector. The cross-modal spatial attention can be formularized in the following steps:
\begin{align}
    M_q &= \text{tanh } ((W_mM + b_m) \oplus W_qq) \label{eq:1} \\ 
    M_i &= \text{softmax } (W_iM_q + b_i) \label{eq:2} \\
    M_{ns} &= \text{Nucleus 
    Sampling } (M_i) \label{eq:3} \\
    M_o &= M_{ns} \otimes M, \label{eq:4}
\end{align}
\noindent in which $W_m \in \mathbb{R}^{k \times C}$, $W_q \in \mathbb{R}^{k \times d}$, and $W_i \in \mathbb{R}^{k}$ are linear transformation weights, and $b_m \in \mathbb{R}^{k}$ and $b_i \in \mathbb{R}^1$ are biases, where $k$ is a pre-defined hyper-parameter, and $\oplus$ represents the broadcast addition operation of a tensor and a vector. In Eq.\ref{eq:2}, the softmax function is applied to the ($H \times W$) dimensions, and $M_i \in \mathbb{R}^{S \times H \times W}$ is the tensor of the spatial attention scores corresponding to each feature map. Finally, to make the input feature maps $M$ sparse while reserving important information, firstly, we perform Nucleus Sampling~\citep{holtzman2019curious} on $M_i$ to get the top-$p$ portion of the probability mass in each attention score map ($p$ is a pre-defined hyper-parameter in the range of $(0, 1]$). In $M_{ns}$, the points selected by the Nucleus Sampling are set to one and the others are set to zero. Then, we do broadcast point-wise multiplication between $M_{ns}$ and $M$ to generate the output $M_o$. Therefore, $M_o$ is a sparse tensor with some positions being zero, and the degree of sparsity is controlled by $p$.


\subsubsection{Sparse CNN}
We use the submanifold sparse CNN~\citep{SubmanifoldSparseConvNet} after the cross-modal attention layer. 
It is leveraged for processing low-dimensional data which lies in a space of higher dimensionality. In the multimodal emotion recognition task, we assume that only part of the data is related to the recognition of emotions (an intuitive example is given in Figure~\ref{fig:intro}), which makes it align with the sparse setting. In our model, the sparse CNN layer accepts the output from the cross-modal attention layer, and does convolution computation only at the active positions. Theoretically, in terms of the amount of computation (FLOPs) at a single location, a standard convolution costs $z^2mn$ FLOPs, and a sparse convolution costs $amn$ FLOPs, where $z$ is the kernel size, $m$ is the number of input channels, $n$ is the number of output channels, and $a$ is the number of active points at this location. Therefore, considering all locations and all layers, the sparse CNN can help to significantly reduce computation.

\section{Experiments}

\subsection{Evaluation Metrics} \label{eval_metrics}
Following prior works~\citep{tsai2018learning,wang2019words,tsai2019multimodal,Dai2020ModalityTransferableEE}, we use the accuracy and F1-score to evaluate the models on the IEMOCAP dataset. On the CMU-MOSEI dataset, we use the weighted accuracy instead of the standard accuracy. Additionally, according to \citet{Dai2020ModalityTransferableEE}, we use the standard binary F1 rather than the weighted version.

\paragraph{Weighted Accuracy} Similar to existing works~\citep{zadeh2018multimodal,akhtar2019multi}, we use the weighted accuracy (WAcc)~\citep{tong2017combating} to evaluate the CMU-MOSEI dataset, which contains many more negative samples than positive ones on each emotion category. If normal accuracy is used, a model will still get a fine score when predicting all samples to be negative. The formula of the weighted accuracy is
\begin{align*}
    \text{WAcc.} = \frac{TP \times N/P + TN}{2N} ,
\end{align*}
\noindent in which P means total positive, TP true positive, N total negative, and TN true negative.


\begin{table*}[t]
\centering
\resizebox{\textwidth}{!}{
\begin{tabular}{lc|cccccccccccccc}
\toprule
\multirow{2}{*}{\textbf{Model}} & \multirow{2}{*}{\textbf{\begin{tabular}[c]{@{}c@{}}\#FLOPs\\ ($\times 10^9$) \end{tabular}}} & \multicolumn{2}{c}{\textbf{Angry}} & \multicolumn{2}{c}{\textbf{Excited}} & \multicolumn{2}{c}{\textbf{Frustrated}} & \multicolumn{2}{c}{\textbf{Happy}} & \multicolumn{2}{c}{\textbf{Neutral}} & \multicolumn{2}{c}{\textbf{Sad}} & \multicolumn{2}{c}{\textbf{Average}} \\
 &  & Acc. & F1 & Acc. & F1 & Acc. & F1 & Acc. & F1 & Acc. & F1 & Acc. & F1 & Acc. & F1 \\ 
\midrule
LF-LSTM & - & 71.2 & 49.4 & 79.3 & 57.2 & 68.2 & 51.5 & 67.2 & 37.6 & 66.5 & 47.0 & 78.2 & 54.0 & 71.8 & 49.5 \\
LF-TRANS & - & 81.9 & 50.7 & 85.3 & 57.3 & 60.5 & 49.3 & 85.2 & 37.6 & 72.4 & 49.7 & 87.4 & 57.4 & 78.8 & 50.3 \\
EmoEmbs$^\dagger$ & - & 65.9 & 48.9 & 73.5 & 58.3 & 68.5 & 52.0 & 69.6 & 38.3 & 73.6 & 48.7 & 80.8 & 53.0 & 72.0 & 49.8 \\
MulT$^\dagger$ & - & 77.9 & 60.7 & 76.9 & 58.0 & 72.4 & 57.0 & 80.0 & 46.8 & 74.9 & 53.7 & 83.5 & 65.4 & 77.6 & 56.9 \\
\midrule
FE2E & 8.65 & \textbf{88.7} & \textbf{63.9} & \textbf{89.1} & \textbf{61.9} & 71.2 & 57.8 & \textbf{90.0} & 44.8 & \textbf{79.1} & \textbf{58.4} & \textbf{89.1} & \textbf{65.7} & \textbf{84.5} & \textbf{58.8} \\
MESM ($p=0.7$) & \textbf{5.18} & 88.2 & 62.8 & 88.3 & 61.2 & \textbf{74.9} & \textbf{58.4} & 89.5 & \textbf{47.3} & 77.0 & 52.0 & 88.6 & 62.2 & \textbf{84.4} & 57.4 \\ 
\bottomrule
\end{tabular}
}
\caption{The results on the IEMOCAP dataset. \#FLOPs is the number of floating point operations per second. We report the accuracy (Acc.) and the F1-score on six emotion categories: \textit{angry}, \textit{excited}, \textit{frustrated}, \textit{happy}, \textit{neutral} and \textit{sad}. We re-run the models marked by $^\dagger$, as we use two more categories and the split is different.}
\label{tab:iemocap}
\end{table*}

\begin{table*}[t]
\centering
\resizebox{\textwidth}{!}{
\begin{tabular}{lc|cccccccccccccc}
\toprule
\multirow{2}{*}{\textbf{Model}} & \multirow{2}{*}{\textbf{\begin{tabular}[c]{@{}c@{}}\#FLOPs\\ ($\times 10^9$) \end{tabular}}} & \multicolumn{2}{c}{\textbf{Angry}} & \multicolumn{2}{c}{\textbf{Disgusted}} & \multicolumn{2}{c}{\textbf{Fear}} & \multicolumn{2}{c}{\textbf{Happy}} & \multicolumn{2}{c}{\textbf{Sad}} & \multicolumn{2}{c}{\textbf{Surprised}} & \multicolumn{2}{c}{\textbf{Average}} \\
 &  & WAcc. & F1 & WAcc. & F1 & WAcc. & F1 & WAcc. & F1 & WAcc. & F1 & WAcc. & F1 & WAcc. & F1 \\ 
\midrule
LF-LSTM & - & 64.5 & 47.1 & 70.5 & 49.8 & 61.7 & 22.2 & 61.3 & 73.2 & 63.4 & 47.2 & 57.1 & 20.6 & 63.1 & 43.3 \\
LF-TRANS & - & 65.3 & 47.7 & 74.4 & 51.9 & 62.1 & 24.0 & 60.6 & 72.9 & 60.1 & 45.5 & 62.1 & 24.2 & 64.1 & 44.4 \\
EmoEmbs$^\dagger$ & - & 66.8 & 49.4 & 69.6 & 48.7 & 63.8 & 23.4 & 61.2 & 71.9 & 60.5 & 47.5 & 63.3 & 24.0 & 64.2 & 44.2 \\
MulT$^\dagger$ & - & 64.9 & 47.5 & 71.6 & 49.3 & 62.9 & 25.3 & \textbf{67.2} & \textbf{75.4} & 64.0 & 48.3 & 61.4 & 25.6 & 65.4 & 45.2 \\
\midrule
FE2E & 8.65 &\textbf{67.0} & \textbf{49.6} & \textbf{77.7} & \textbf{57.1} & 63.8 & 26.8 & 65.4 & 72.6 & \textbf{65.2} & \textbf{49.0} & \textbf{66.7} & \textbf{29.1} & \textbf{67.6} & \textbf{47.4} \\
MESM (0.5) & \textbf{4.34} & 66.8 & 49.3 & 75.6 & 56.4 & \textbf{65.8} & \textbf{28.9} & 64.1 & 72.3 & 63.0 & 46.6 & 65.7 & 27.2 & 66.8 & 46.8 \\ 
\bottomrule
\end{tabular}
}
\caption{The results on the CMU-MOSEI dataset. WAcc stands for weighted accuracy. We report the accuracy and the F1-score on six emotion categories: \textit{angry}, \textit{disgusted}, \textit{fear}, \textit{happy}, \textit{sad} and \textit{surprised}. We re-run the models marked by $^\dagger$, as the data we use is unaligned along the sequence length dimension and the split is different.}
\label{tab:mosei}
\end{table*}

\subsection{Baselines} \label{baseline}
For our baselines, we use a two-phase pipeline, which consists of a feature extraction step and an end-to-end learning step.

\paragraph{Feature Extraction} We follow the feature extraction procedure in the previous works~\citep{zadeh2018multimodal,tsai2018learning,tsai2019multimodal,rahman2020integrating}. For the visual data, we extract 35 facial action units (FAUs) using the OpenFace library\footnote{\url{https://github.com/TadasBaltrusaitis/OpenFace}}~\cite{tadas2015openface,baltrusaitis2018openface} for the image frames in the video, which capture the movement of facial muscles~\citep{ekman1980facial}. For the acoustic data, we extract a total of 142 dimension features consisting of 22 dimension bark band energy (BBE) features, 12 dimension mel-frequency cepstral coefficient (MFCC) features, and 108 statistical features from 18 phonological classes. We extract the features per 400 ms time frame using the DisVoice library\footnote{\url{https://github.com/jcvasquezc/DisVoice}}~\citep{vsquez2018articulation,vsquez2019phonet}. For textual data, we use the pre-trained GloVe~\citep{pennington2014glove} word embeddings (glove.840B.300d\footnote{\url{https://nlp.stanford.edu/projects/glove/}}).

\paragraph{Multimodal Learning} As different modalities are unaligned in the data, we cannot compare our method with existing works that can only handle aligned input data. We use four multimodal learning models as baselines: the late fusion LSTM (LF-LSTM) model, the late fusion Transformer (LF-TRANS) model, the Emotion Embeddings (EmoEmbs) model~\cite{Dai2020ModalityTransferableEE}, and the Multimodal Transformer (MulT) model~\cite{tsai2019multimodal}.
They receive the hand-crafted features extracted from the first step as input and give the classification decisions. 

\subsection{Training Details}
We use the Adam optimizer~\citep{kingma2014adam} for the training of every model we use. For the loss function, we use the binary cross-entropy loss as both of the datasets are multi-class and multi-labelled. In addition, the loss for the positive samples is weighted by the ratio of the number of positive and negative samples to mitigate the imbalance problem. For all of the models, we perform an exhaustive hyper-parameter search to ensure we have solid comparisons. The best hyper-parameters are reported in Appendix A. Our experiments are run on an Nvidia 1080Ti GPU, and our code is implemented in the PyTorch~\citep{paszke2019pytorch} framework v1.6.0. We perform preprocessing for the text and audio modalities. For the text modality, we perform word tokenization for our baseline and subword tokenization for our end-to-end model. We limit the length of the text to up to 50 tokens. For the audio modality, we use mel-spectrograms with a window size of 25 ms and stride of 12.5 ms and then chunk the spectrograms per 400 ms time window. 


\begin{figure*}[!t]
    \centering
    \includegraphics[width=\linewidth]{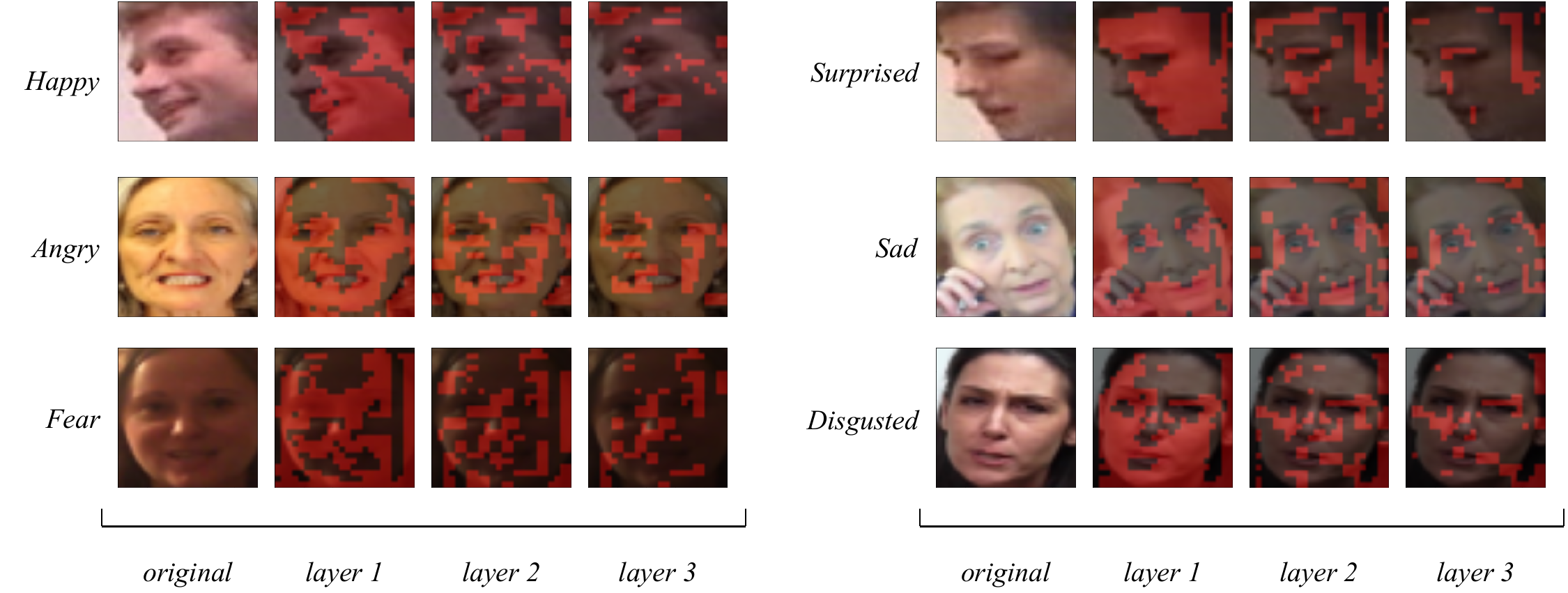}
    \caption{Case study of MESM on six basic emotion categories (happy, sad, angry, surprised, fear, disgusted). From left to right, we show the original image and the Nucleus Sampling ($p=0.6$) result over points in each attention layer. Red regions represent the points that are computed for the next layer.}
    \label{fig:case_study}
\end{figure*}

\section{Analysis}
\subsection{Results Analysis}
In Table~\ref{tab:iemocap}, we show the results on the IEMOCAP dataset. Compared to the baselines, the fully end-to-end (FE2E) model surpasses them by a large margin on all the evaluation metrics. Empirically, this shows the superiority of the FE2E model over the two-phase pipeline. Furthermore, our MESM achieves comparable results with the FE2E model, while requiring much less computation in the feature extraction. Here, we only show the results of MESM with the best $p$ value of the Nucleus Sampling. In Section~\ref{ns}, we conduct a more detailed discussion of the effects of the top-p values. We further evaluate the methods on the CMU-MOSEI dataset and the results are shown in Table~\ref{tab:mosei}. We observe similar trends on this dataset.

\subsection{Case Study}

\begin{figure}[t]
    \centering
    \includegraphics[width=0.85\linewidth]{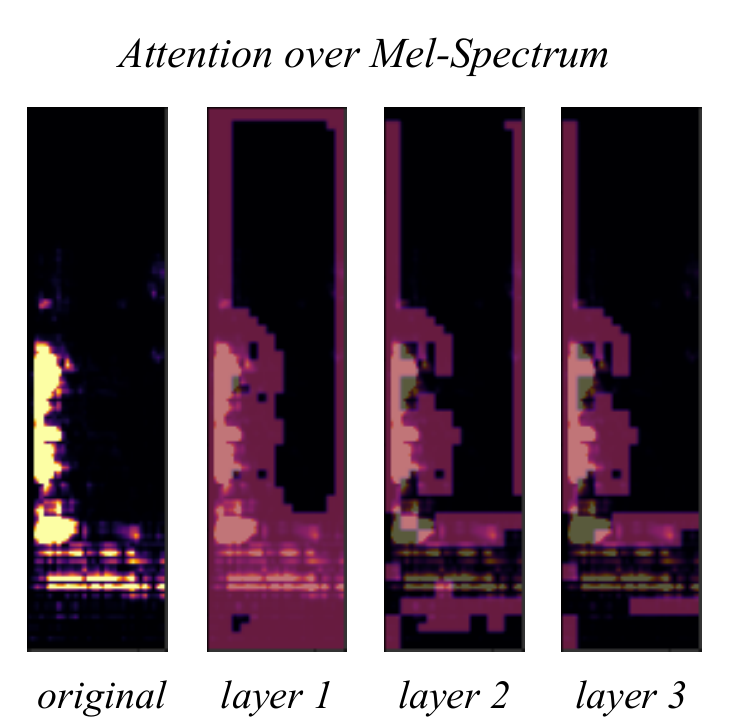}
    \caption{Visualization of cross-modal attention of the acoustic modality. We only show the highest 10\% value of mel-spectrogram in each image. From left to right, we show the original image and the Nucleus Sampling ($p=0.6$) results over points in each attention layer. Red regions represent the active points that will go to the next sparse CNN layer.}
    \label{fig:audio_case_study}
\end{figure}

To improve the interpretability and gain more insights from our model, we visualize the attention maps of our sparse cross-modal attention mechanism on the six basic emotions: happy, sad, angry, surprised, fear, and disgusted. As shown in Figure~\ref{fig:case_study}, in general, the models attend to several regions of interest such as the mouth, eyes, eyebrows, and facial muscles between the mouth and the eyes. We verify our method by comparing the regions that our model captures based on the facial action coding system (FACS)~\citep{ekman1997face}. Following the mapping of FACS to human emotion categories~\citep{basori2016facial,ahn2017insideout}, we conduct empirical analysis to validate the sparse cross-modal attention on each emotion category. For example, the emotion \textit{happy} is highly influenced by raising of the lip on both ends, while \textit{sad} is related to a lowered lip on both ends and downward movement of the eyelids. \textit{Angry} is determined from a narrowed gap between the eyes and thinned lips, while \textit{surprised} is expressed with an open mouth and raising of the eyebrows and eyelids. \textit{Fear} is indicated by a rise of the eyebrows and upper eyelids, and also an open mouth with the ends of the lips slightly moving toward the cheeks. For the emotion \textit{disgusted}, wrinkles near the nose area and movement of the upper lip region are the determinants.

Based on the visualization of the attention maps on the visual data in Figure~\ref{fig:case_study}, the MESM can capture most of the specified regions of interest for the six emotion categories. For the emotion \textit{angry}, the sparse cross-modal attention can retrieve the features from the lip region quite well, but it sometimes fails to capture the gap between the eyes. For \textit{surprised}, the eyelids and mouth regions can be successfully captured by MESM, but sometimes the model fails to consider the eyebrow regions. 
For the acoustic modality, it is hard to analyse the attention in terms of emotion labels. We show a general visualization of the attention maps over the audio data in Figure~\ref{fig:audio_case_study}. The model attends to the regions with high spectrum values in the early attention layer, and more points are filtered out after going through further cross-modal attention layers. More visualized examples are provided in Appendix B.

\subsection{Effects of Nucleus Sampling} \label{ns}

\begin{figure}[!t]
  \centering
  \includegraphics[width=\linewidth]{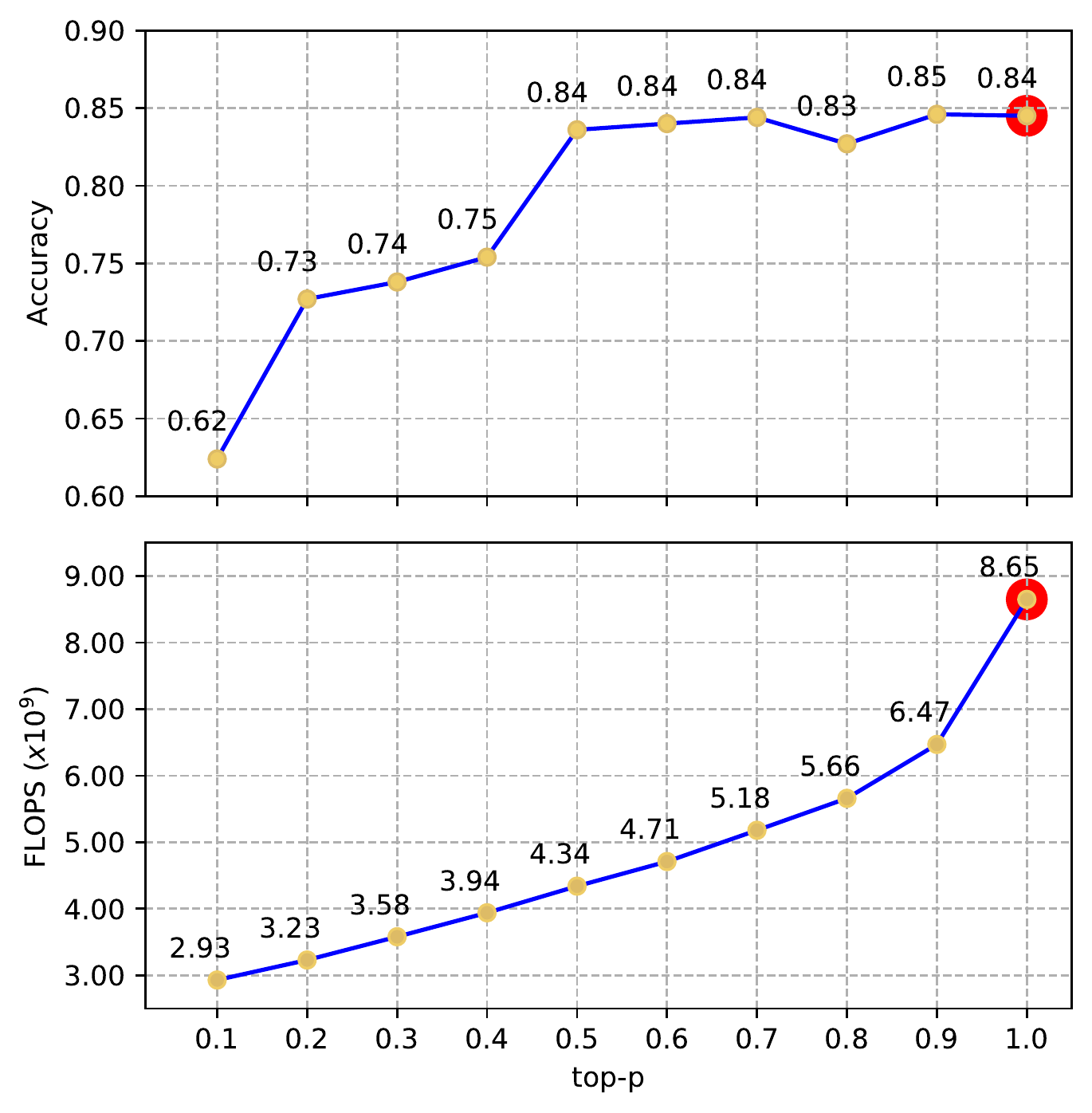}
\caption{The trend line of the \textbf{Top:} Weighted Accuracy and \textbf{Bottom:} FLOPs (x$10^9$)) of the MESM with different top-p values used in the Nucleus Sampling. \includegraphics[height=\fontcharht\font`\B]{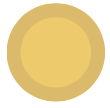} represents performance of MESM, while \includegraphics[height=\fontcharht\font`\B]{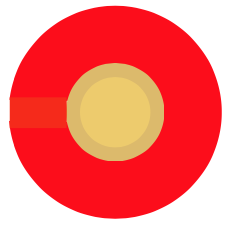} represents performance of the FE2E model}

\label{fig:ns}
\end{figure}

To have an in-depth understanding of the effects of Nucleus Sampling on the MESM, we perform more experiments with different top-p values ranging from 0 to 1, with a step of 0.1. As shown in Figure~\ref{fig:ns}, empirically, the amount of computation is reduced consistently with the decrease of the top-p values. In terms of performance, with a top-p value from 0.9 to 0.5, there is no significant drop in the evaluation performance. Starting from 0.5 to 0.1, we can see a clear downgrade in the performance, which means some of the useful information for recognizing the emotion is excluded. The inflection point of this elbow shaped trend line can be an indicator to help us make a decision on the value of the top-$p$. Specifically, with a top-$p$ of 0.5, the MESM can achieve comparable performance to the FE2E model with around half of the FLOPs in the feature extraction.


\begin{table}[!h]
\centering
\resizebox{0.42\textwidth}{!}{
\begin{tabular}{lccc}
\toprule
\textbf{Model} & \textbf{Mods.} & \textbf{Avg. Acc} & \textbf{Avg. F1} \\ \midrule
\multirow{7}{*}{FE2E} & TAV & 84.5 & 58.5 \\
 & TA & 83.7 & 54.0 \\
 & TV & 82.8 & 55.7 \\
 & VA & 81.2 & 54.4 \\
 & T & 80.8 & 50.0 \\
 & A & 73.3 & 44.9 \\
 & V & 78.2 & 49.8 \\ \midrule
\multirow{3}{*}{MESM} & TAV & 84.4 & 57.3 \\
 & TA & 83.6 & 56.7 \\
 & TV & 82.1 & 56.0 \\ \bottomrule
\end{tabular}
}
\caption{Results of the ablation study of our fully end-to-end model (FE2E) and multimodal end-to-end sparse model (MESM) on the IEMOCAP dataset. In the Mods. (modalities) column, the T/A/V indicates the existence of the textual (T), acoustic (A), and visual (V) modalities.}
\label{tab:ablation}
\end{table}

\section{Ablation Study}
We conduct a comprehensive ablation study to further investigate how the models perform when one or more modalities are absent. The results are shown in Table~\ref{tab:ablation}. Firstly, we observe that the more modalities the more improvement in the performance. TAV, representing the presence of all three modalities, results in the best performance for both models, which shows the effectiveness of having more modalities. Secondly, with only a single modality, the textual modality results in better performance than the other two, which is similar to the results of previous multimodal works. This phenomenon further validates that using textual (T) to attend to acoustic (A) and visual (V) in our cross-modal attention mechanism is a reasonable choice. Finally, with two modalities, the MESM can still achieve a performance that is on par with the FE2E model or is even slightly better.

\section{Conclusion and Future Work}
In this paper, we first compare and contrast the two-phase pipeline and the fully end-to-end (FE2E) modelling of the multimodal emotion recognition task. Then, we propose our novel multimodal end-to-end sparse model (MESM) to reduce the computational overhead brought by the fully end-to-end model. Additionally, we reorganize two existing datasets to enable fully end-to-end training. The empirical results demonstrate that the FE2E model has an advantage in feature learning and surpasses the current state-of-the-art models that are based on the two-phase pipeline. Furthermore, MESM is able to halve the amount of computation in the feature extraction part compared to FE2E, while maintaining its performance.
In our case study, we provide a visualization of the cross-modal attention maps on both visual and acoustic data. It shows that our method can be interpretable, and the cross-modal attention can successfully select important feature points based on different emotion categories. For future work, we believe that incorporating more modalities into the sparse cross-modal attention mechanism is worth exploring since it could potentially enhance the robustness of the sparsity (selection of features).

\section*{Acknowledgement}
This work is funded by MRP/055/18 of the Innovation Technology Commission, the Hong Kong SAR Government.

\bibliography{anthology,custom}
\bibliographystyle{acl_natbib}

\newpage
\appendix

\section{Hyper-parameter Settings}
\label{sec:appendix_a}
\begin{table}[h]
\centering
\resizebox{\linewidth}{!}{
\begin{tabular}{lcccc}
\toprule
 & \multicolumn{2}{c}{IEMOCAP} & \multicolumn{2}{c}{CMU-MOSEI} \\
 & FE2E & MESM & FE2E & MESM \\ \midrule
Batch size & 8 & 8 & 8 & 8 \\
Learning rate & 5e-5 & 5e-5 & 5e-5 & 5e-5 \\
Dim & 64 & 64 & 64 & 64 \\
\#Heads & 4 & 4 & 4 & 4 \\
\#Layers & 4 & 4 & 4 & 4 \\
Max text len & 50 & 100 & 50 & 100 \\
N & - & 3 & - & 3 \\ \bottomrule
\end{tabular}
}
\caption{The best hyper-parameters used in training for the two datasets.}
\label{tab:hyperparams}
\end{table}

\section{Case Study on Acoustic Modality}
\begin{figure*}[!h]
    \centering
    \includegraphics[width=\linewidth]{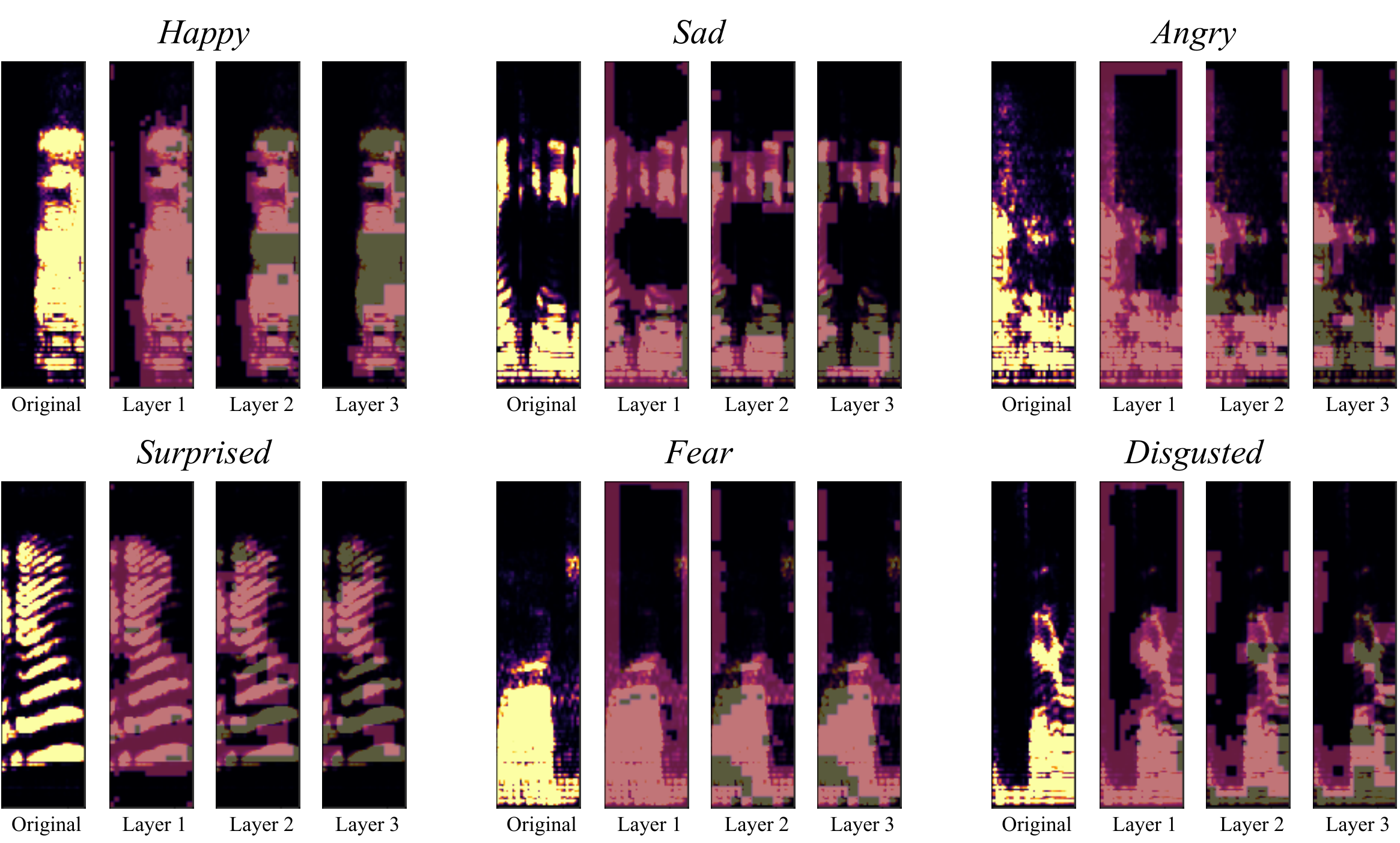}
    \caption{Case study of the sparse cross-modal attention maps on six basic emotion categories (\textit{happy}, \textit{sad}, \textit{angry}, \textit{surprised}, \textit{fear}, \textit{disgusted}) on the audio modality. From the left to right, we show the original image and the Nucleus Sampling results over feature points in each attention layer. Red regions represent the active points that will be computed in the next sparse layer.}
    \label{fig:app_aud_case_study}
\end{figure*}

\label{sec:appendix_b}
We provide more visualized examples of the sparse cross-modal attention maps of the acoustic modality in Figure~\ref{fig:app_aud_case_study}.

\end{document}